\documentclass[10pt,a4paper]{article}

\usepackage[final]{lrec2026} 
\usepackage{graphicx}
\usepackage{float}
\usepackage{url}
\usepackage{natbib}
\usepackage{caption}
\usepackage{algorithm}
\usepackage{algorithmic}
\usepackage{newfloat}
\usepackage{listings}
\usepackage{booktabs}

\DeclareCaptionStyle{ruled}{labelfont=normalfont,labelsep=colon,strut=off}
\floatstyle{ruled}
\newfloat{listing}{tb}{lst}{}
\floatname{listing}{Listing}

\title{Can Large Language Models Imitate Human Speech for Clinical Assessment? LLM-Driven Data Augmentation for Cognitive Score Prediction}

\name{Si-Belkacem Yamine Ketir$^1$, Lenard Paulo Tamayo$^2$,\\
Shohei Hisada$^2$, Shaowen Peng$^2$, Shoko Wakamiya$^2$, Eiji Aramaki$^2$}

\address{
    $^1$ Télécom SudParis, France \\
    $^2$ Nara Institute of Science and Technology, Japan \\[4pt]
    {\small si-belkacem.ketir@telecom-sudparis.eu} \\
    {\small \{lenard\_paulo.tamayo.ly4, peng.shaowen\}@naist.ac.jp} \\
    {\small \{s-hisada, wakamiya, aramaki\}@is.naist.jp}
}

\abstract{
Accurate assessment of cognitive decline from spontaneous speech remains challenging due to limited dataset size and class imbalance. In this work, we propose a large language model (LLM)-driven data augmentation framework to improve the prediction of  cognitive scores from speech. Experiments are conducted on a Japanese corpus in which each participant provides both a spontaneous oral narrative and a written response to the same clinical prompt. The written responses serve as semantic anchors to generate multiple oral-like monologues in different styles using GPT-5. We then predict Hasegawa Dementia Scale scores, a widely used cognitive screening tool in Japan, using a Partial Least Squares regression model trained on Sentence-BERT speech embeddings. We investigate two augmentation strategies: random class-balanced selection, which yields moderate but unstable improvements, and similarity-guided class-balanced selection. The latter prioritizes semantically close synthetic samples, leading to more consistent improvements and substantially reducing prediction error for minority low-score participants while maintaining performance for the majority group. Overall, our findings demonstrate the potential of semantically guided LLM-driven augmentation as a principled approach for addressing class imbalance and improving data efficiency in clinical speech analysis.
\\ \newline
}

\begin{document}

\maketitleabstract
\section{Introduction}

The global increase in life expectancy has made dementia one of the most pressing public health challenges of the 21st century, with cases expected to triple by 2050 \citep{Livingston2020Lancet, WHO2019DementiaGuidelines}. In the absence of curative treatments, early detection of cognitive decline is critical for enabling timely interventions. 
Although standardized neuropsychological assessments, such as the Mini-Mental State Examination (MMSE) and the Revised Hasegawa Dementia Scale (HDS-R; hereafter referred to as HDS) \citep{maeshima2024neuropsychological}, remain the gold standard for cognitive evaluation, they possess inherent limitations. 
These tests require administration by trained professionals and can impose significant psychological stress on elderly patients, potentially confounding the results. 
Furthermore, as they are typically conducted only after symptoms become clinically apparent, their sensitivity to subtle, subclinical early-stage decline is limited. Moreover, frequent longitudinal monitoring is often impractical due to practice effects, where repetitive exposure to specific test items can artificially inflate scores, masking the true cognitive trajectory
\citep{maeshima2024neuropsychological, Tiberti1998, Igarashi2022}.

\begin{figure}[t]
\begin{center}
    \includegraphics[width=1.0\linewidth]{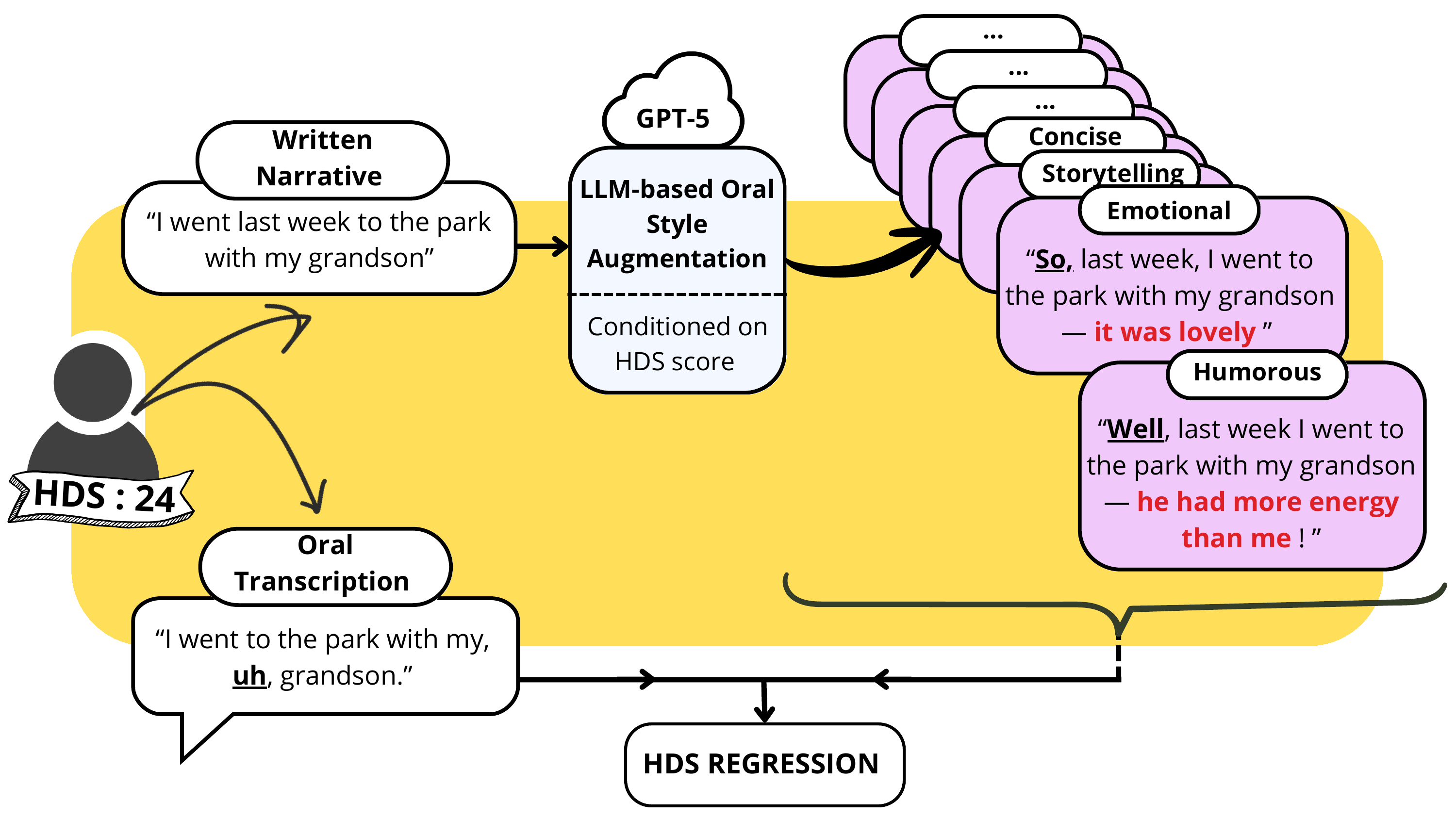}
\caption{Overview of the proposed LLM-driven data augmentation framework for cognitive score prediction from speech. Underlined terms indicate oral markers, and terms in red indicate stylistic features.}
    \label{fig:overview0}
\end{center}
\end{figure}

To overcome these limitations, spontaneous speech analysis has emerged as a non-invasive and cost-effective biomarker for cognitive health \citep{lima2025spokenbiomarker}. Language, as a complex cognitive task integrating memory, executive function, and semantic processing, exhibits subtle changes in syntax, vocabulary, and fluency that often precede measurable declines in other domains \citep{szatloczki2015speaking, hoffmann2010temporal}. This makes speech a promising signal for early detection.

Despite its potential, developing robust machine learning models for speech-based cognitive assessment faces several challenges. Clinical datasets remain small due to ethical and logistical constraints \citep{qi2023noninvasive}, and subsets such as the Pitt Corpus exhibit class imbalances \citep{jahan2024_early_dementia}. Standard lexical data augmentation techniques, such as synonym replacement, random insertion, swap, or deletion, have been applied to cognitive assessment tasks \citep{Igarashi2022}. However, when the objective is to model subtle linguistic markers of dementia, such as impaired coherence, syntactic irregularities, or word-finding difficulties, these generic transformations may inadvertently modify clinically meaningful signals. Moreover, arbitrary lexical or syntactic modifications may introduce artifacts unrelated to genuine cognitive decline, reducing clinical validity.

To address these challenges, we propose a clinically-guided synthetic data generation framework powered by a large language model. Unlike generic lexical augmentation strategies, our approach leverages the particular setting in which each patient provides both written and oral responses to a standardized cognitive prompt.
We adopt a cross-modal-inspired approach by conditioning an LLM on structured written narratives  to generate synthetic oral-style transcriptions. This conditioning mechanism enables controlled variation: the generated samples preserve factual and cognitive content while introducing spontaneous speech characteristics such as disfluencies, reduced syntactic complexity, and stylistic variability. As illustrated in Figure~\ref{fig:overview0}, this approach mitigates hallucination risks and preserves the clinical validity of subtle speech patterns.

The main contributions of this paper are as follows:
\begin{itemize}
    \item An LLM-based cross-modal-inspired data augmentation framework that transforms written narratives into oral-style speech while preserving dementia-specific linguistic markers and addressing data scarcity.
    \item Demonstration that similarity-guided filtering is essential for maintaining synthetic data quality and fidelity to authentic speech patterns.
    \item Validation of this approach on a cognitive score regression task using a small and imbalanced clinical dataset.
\end{itemize}

This work provides a principled approach for responsible synthetic data generation in clinical speech analysis and demonstrates the potential of large language models for data-efficient cognitive assessment.

\section{Related Work}

Automatic assessment of cognitive decline from spontaneous speech has attracted growing interest in both computational linguistics and clinical AI. Early studies primarily focus on binary classification of dementia, often relying on publicly available speech corpora such as DementiaBank \citep{fraser2016linguistic}, which also inspired standardized datasets like the ADReSS challenge corpus \citep{zolnour2025llmcare}. Despite their widespread use, these datasets are limited in size and some suffer from class imbalance. For instance, the Pitt corpus includes 955 recordings from healthy controls versus 586 from dementia patients, posing challenges for building robust and generalizable models \citep{jahan2024_early_dementia, hledikova2022data, qi2023noninvasive}.

Early approaches relied on handcrafted acoustic and linguistic features, such as pause duration, speech rate, lexical richness, syntactic complexity, and disfluency markers, which were then used with traditional machine learning models to discriminate between cognitively healthy subjects and patients with dementia \citep{fraser2016linguistic, toth2015automatic}. 

With the advent of deep learning, more recent studies have shifted toward representation learning approaches based on pretrained language and speech models. Contextual embeddings derived from models such as BERT have been shown to capture clinically relevant linguistic patterns and can be effectively used for dementia detection and cognitive score prediction \citep{balagopalan2020to}. In parallel, many approaches have adopted multimodal strategies, combining acoustic and linguistic features through fusion mechanisms to improve classification performance \citep{hledikova2022data, zolnour2025llmcare}. These multimodal methods leverage complementary information from different data sources, but they generally focus on feature-level or late fusion.

To overcome limitations imposed by small datasets, several studies have explored data augmentation techniques. These include text-based augmentations, such as synonym substitution, paraphrasing, or random sentence modification, as well as audio-based augmentations, including noise addition, pitch and time shifting, and time/frequency masking \citep{hledikova2022data}. Recent work \citep{Igarashi2022} demonstrated that applying text-based augmentation to a small dataset of Japanese older adults improved classification performance, highlighting the potential of such methods to enhance cognitive assessment accuracy. More recently, large language models (LLMs) have been investigated as generators of synthetic clinical text to further improve data diversity and model robustness \citep{hledikova2022data, liu2025generating}.

Despite these advances, most prior studies focus on binary classification of dementia versus non-dementia, with few attempting to predict actual cognitive scores. Moreover, no work has explored LLM-driven cross-modal augmentation for cognitive score prediction from spontaneous speech, nor the use of written patient narratives to generate oral-style monologues. These gaps motivate methods capable of handling small datasets and predicting continuous cognitive scores using cross-modal, LLM-based augmentation strategies.

\section{Task Formulation and Dataset}
\label{sec:3}
\subsection{Task Formulation}

The objective of this study is to predict cognitive scores from speech data using a regression approach, with the ultimate goal of supporting early detection of cognitive decline. In our setting, each participant provides a spontaneous narrative in response to a single standardized cognitive prompt, and the target of prediction is the participant’s HDS score, a widely used cognitive screening measure in Japan.

While the primary input for prediction is the oral transcription of the participant’s narrative, the handwritten responses to the same prompt are leveraged as auxiliary data during training only. These written texts are not used during inference but serve to enrich the training set via synthetic data generation. This strategy allows the model to better capture subtle linguistic markers of cognitive decline despite the limited size of the available speech dataset.

\subsection{Dataset Overview}

This study relies on a subset of the Japanese elderly speech corpus with healthy controls (GSK2018-A)\footnote{\url{https://www.gsk.or.jp/catalog/gsk2018-a}(in Japanese)} distributed by the Gengo Shigen Kyokai (GSK). The subset used in this study contains speech data from 30 elderly participants aged 72 to 86 years, each providing a spontaneous narrative in response to the prompt: \textit{``Please tell us about the last good thing that happened to you.''} Along with high-quality audio recordings, the dataset includes handwritten responses to the same prompt as a controlled linguistic reference, and a clinical annotation in the form of the HDS score. This design elicits free-form autobiographical speech while maintaining thematic consistency across participants, making the corpus well suited for linguistic and paralinguistic analysis.

\subsection{Task Challenges}

\begin{figure}[tb]
\centering
\includegraphics[width=\linewidth]{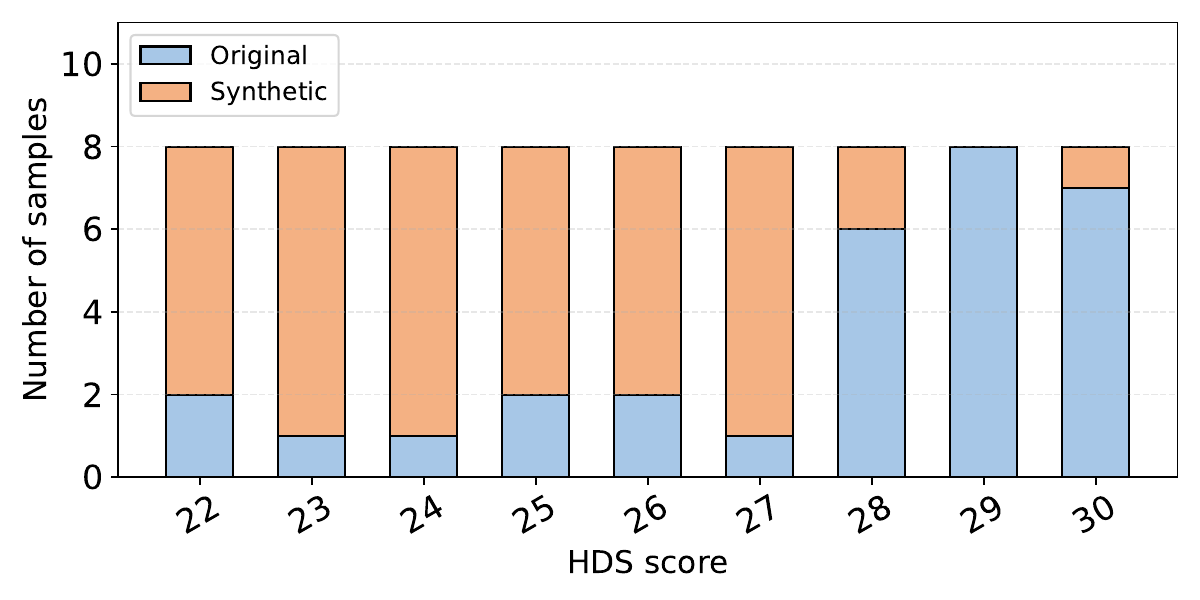}
\caption{Distribution of HDS score classes before and after synthetic data augmentation. Blue ("original") refers to manually annotated oral transcriptions, while orange ("synthetic") refers to synthetic data generated from patients' written responses.}
\label{fig:dataset_balance}
\end{figure}

Despite the richness of the dataset, the task presents significant challenges. Manually annotated oral transcriptions are only available for participants aged 74 years and above, and automatic transcriptions generated using the Japanese version of OpenAI Whisper-1 were required for younger participants, introducing potential variability in transcription quality. In addition, analysis of the HDS score distribution in the original dataset (represented by the blue bars in Figure~\ref{fig:dataset_balance}) reveals a strong imbalance: scores predominantly range between 28 and 30, corresponding to cognitively healthy states, while lower scores (22–27) are under-represented. This scarcity of low-score data complicates regression modeling and motivates the need for data augmentation strategies that can enhance model performance across the full range of cognitive scores.


\section{Proposed Method}
\label{sec:4}
\subsection{Overview}

\begin{figure}[tb!]
\centering
\includegraphics[width=1.0\linewidth]{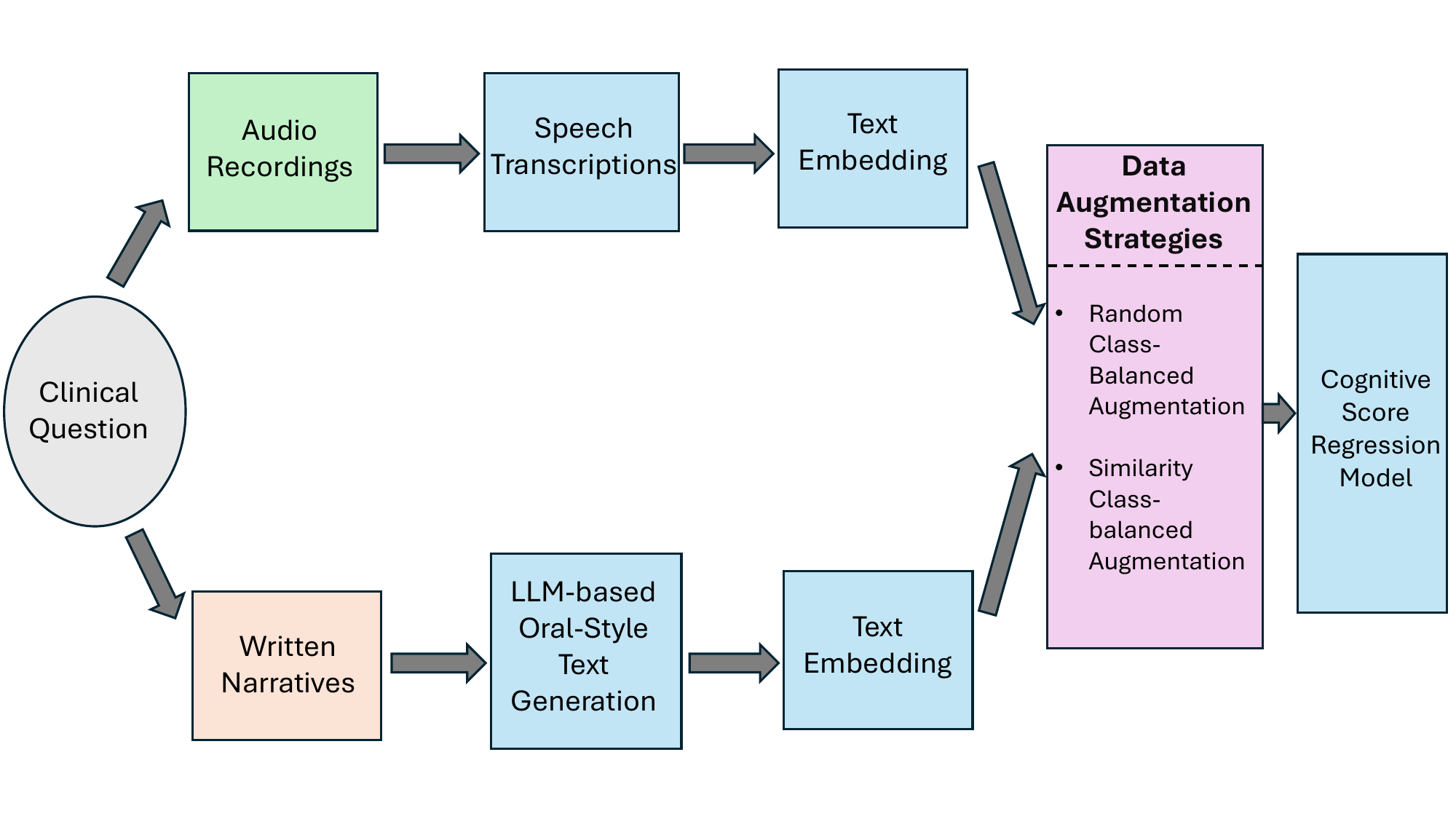}
\caption{LLM-based data augmentation framework for predicting cognitive scores from speech.}
\label{fig:overview}
\end{figure}

As illustrated in Figures~\ref{fig:overview0} and \ref{fig:overview}, we propose a purely natural language processing (NLP)-based framework to augment speech data for HDS score regression. For each patient, multiple synthetic oral-style monologues are generated using a large language model, conditioned on the patient’s written narrative and associated HDS score. Seven synthetic transcriptions are produced per patient, preserving the original semantic content while introducing natural variations in language and discourse.

To reflect cognitive variability, these synthetic monologues incorporate oral markers such as hesitations, fillers, pauses, and simplified expressions, with the frequency and intensity of these markers informed by the HDS score. The augmented training set is formed by combining the synthetic samples with the original oral transcriptions. Two selection strategies, random selection and similarity-guided selection, are applied to control data quality. A concrete step-by-step example illustrating the full transformation
from written narrative to synthetic monologue is provided in
Appendix~\ref{app:example}.


\subsection{Semantic Anchoring}
Unconstrained text generation can introduce modifications to the original semantic content of patient responses. In a clinical context, maintaining the original meaning of patient data is therefore a necessary constraint for synthetic data generation.

In the GSK2018-A corpus, each patient provides both an oral response and a written narrative answering the same clinical prompt. These two modalities describe the same underlying experience. The written narratives are generally more structured than the oral responses and may contain additional details, as they are produced without time pressure and allow for greater reflection.

To generate high-quality synthetic data, we use GPT-5, which provides the best performance among currently available LLMs for text generation tasks \citep{openai2025gpt5}. Written narratives serve as semantic references for conditioning GPT-5, ensuring that the original meaning is preserved. Stylistic transformations are then applied to produce outputs that resemble spontaneous speech. These transformations include the insertion of oral disfluencies such as hesitations and pauses, whose frequency and severity are conditioned on the patient’s HDS score.

For each patient, seven synthetic transcriptions with distinct stylistic configurations are generated. This number is determined by the class distribution of HDS scores in the dataset: the most frequent class contains eight instances, while the least frequent classes contain a single instance. Generating seven synthetic samples per patient allows all HDS score classes to reach the same frequency during class-balanced selection.

\subsection{Style Conditioning and Label Assignment}

For each written narrative, seven synthetic oral-style transcriptions are generated. 
Each transcription is produced using a distinct stylistic configuration, designed to introduce controlled linguistic variability while avoiding near-duplicate generations\footnote{The exact prompt template and style description strings
are provided in Appendix A.}  from the same written narrative.

Concretely, the seven stylistic configurations correspond to (1) a conversational retelling with a relaxed tone, (2) a more emotional and reflective delivery, (3) a fragmented spoken style with short segments and occasional breaks in rhythm, (4) a chronological storytelling narration, (5) a concise spoken form with short and simplified sentences, (6) a slightly playful or humorous tone, and (7) a structural paraphrase that preserves meaning while reorganizing sentence structure using similar vocabulary. While these configurations differ in surface realization, they are all applied under the same semantic constraints imposed by the written narrative, ensuring that the original meaning is preserved across synthetic variants.

These style choices are designed to capture complementary dimensions of natural spoken language variability. In particular, they reflect (i) social and conversational interaction (e.g., conversational, playful), (ii) emotional and reflective expression (e.g., emotional), and (iii) cognitive and structural variation in speech production (e.g., fragmented, concise, storytelling narration, and paraphrasing-based reorganizations).

Once stylistic diversity is introduced, label assignment is performed. Since the original patient’s HDS score is known, all synthetic transcriptions generated from a given written narrative are assigned the same HDS label as the corresponding original oral response. 

During generation, the language model is explicitly conditioned on this score in order to produce speech patterns compatible with the assigned HDS level. This conditioning is implemented through constraints on the realization of oral markers and linguistic complexity. For HDS scores between 28 and 30, generated speech is fluent and well structured, with relatively richer vocabulary. For scores between 25 and 27, mild hesitations and slightly simplified sentence structures are introduced. For scores between 22 and 24, the generated speech exhibits a slower rhythm, simpler phrasing, occasional hesitations and vaguer expressions. These constraints provide a concrete realization of HDS-aware conditioning while limiting uncontrolled variation across synthetic samples.

Finally, a set of global constraints is enforced across all generated samples. In fact, each synthetic monologue is restricted to a length between 150 and 1300 Japanese characters and must include natural Japanese speech markers. 

\subsection{Filtering}
\label{sec:4.4}

To investigate the trade-off between augmentation diversity and the semantic quality of synthetic samples, we compare two class-balanced filtering strategies.

In the random selection strategy, generated samples are randomly selected to balance the HDS score distributions. This diversity-oriented approach assumes that maximizing linguistic variation is beneficial, even if some samples deviate from the patient’s original semantic characteristics, potentially introducing higher label noise.

In contrast, the similarity-guided strategy prioritizes samples with the highest cosine similarity to the corresponding original oral transcripts during class balancing. This fidelity-oriented approach acts as a quality control mechanism, under the assumption that preserving semantic consistency with the patient’s actual speech is crucial for accurate cognitive assessment.

\section{Experimental Setup}

\subsection{Data Generation Details}

The dataset is constructed from 30 original oral transcriptions, each corresponding to a distinct patient. 
For each patient, seven synthetic oral-style monologues are generated using the procedure described in Section~\ref{sec:4}, resulting in a total of 210 synthetic samples.

The complete pool of available data therefore consists of 240 monologues: 30 original oral transcriptions and 210 synthetic monologues. 
However, to balance the HDS score distribution (targeting 8 samples per class), synthetic monologues are selectively added to the minority classes. 
This process results in approximately 42 augmented samples being combined with the 30 original oral transcriptions to form the final datasets used for training and evaluation.

\subsection{Downstream Task and Models}

The downstream task consists of predicting the HDS score from oral-style transcriptions. 
To this end, each transcription, whether original or synthetic, is encoded into a fixed-dimensional semantic representation and used as input to a regression model.

We employ a Japanese Sentence-BERT (SBERT) model (\texttt{sonoisa/sentence-bert-base-ja-mean-tokens-v2}) to map each transcription to a 768-dimensional, semantically meaningful embedding suitable for regression tasks \cite{reimers2019sentencebert}. 
All transcriptions are processed in their raw form, including hesitation markers and other speech disfluencies, in order to preserve characteristics of natural spoken language. 
The resulting embeddings are standardized prior to regression.

To model the relationship between these embeddings and the HDS scores, we use Partial Least Squares (PLS) regression. 
PLS is particularly well suited to this setting, as it jointly performs dimensionality reduction while maximizing covariance with the target variable, which is critical given the high embedding dimensionality and the limited number of patients (30).

As a preliminary validation, we compared PLS to a baseline combining PCA with Ridge regression using original oral transcriptions only. 
PLS consistently achieved better regression performance and was therefore retained for all subsequent experiments.

\subsection{Evaluation Protocol}

Given the limited dataset size, model performance is evaluated using Leave-One-Out Cross-Validation (LOOCV), in which each patient is successively held out for testing while all remaining patients are used for training.

To ensure a fair evaluation and strictly prevent data leakage, we enforce a patient-level separation between training and testing. 
When a patient is used as the test subject, all synthetic monologues generated from that patient’s written narrative are excluded from the training set. 
Consequently, the model is never exposed during training to synthetic data derived from the test patient.

Model performance is assessed using Mean Absolute Error (MAE), Root Mean Squared Error (RMSE), and the coefficient of determination ($R^2$).

The effect of synthetic data augmentation is evaluated by progressively adding filtered synthetic samples to the training set until each HDS score class reaches the maximum observed class frequency (8 samples), following the filtering strategies described in Section~\ref{sec:4.4}. 

For the random selection strategy, the experiment is repeated 30 times to account for sampling variability, and 95\% confidence intervals are reported.

\subsection{Baselines}

To evaluate the impact of synthetic data augmentation, we compare our proposed approach against the following two baseline settings of increasing complexity.

\paragraph{Baseline 1: No augmentation} The first baseline involves no augmentation, with PLS regression trained solely on the original oral transcriptions.
The optimal number of PLS components is determined using nested leave-one-out cross-validation to provide an unbiased estimate of generalization performance.
Across folds, the selected number of components is consistent, with a mean value of approximately 7.

\paragraph{Baseline 2: Gaussian noise augmentation} The second baseline introduces a simple feature-space augmentation strategy, where Gaussian noise ($\sigma = 0.02$) is added to the standardized SBERT embeddings of the original transcriptions.
For this baseline, the number of PLS components is fixed at 7, as determined from the no-augmentation setting, in order to maintain the same model complexity and isolate the effect of data augmentation.

\section{Results and Discussion}

\subsection{Effect of Synthetic Data Augmentation}

\begin{figure}[tb]
\centering
\includegraphics[width=1.03\linewidth]{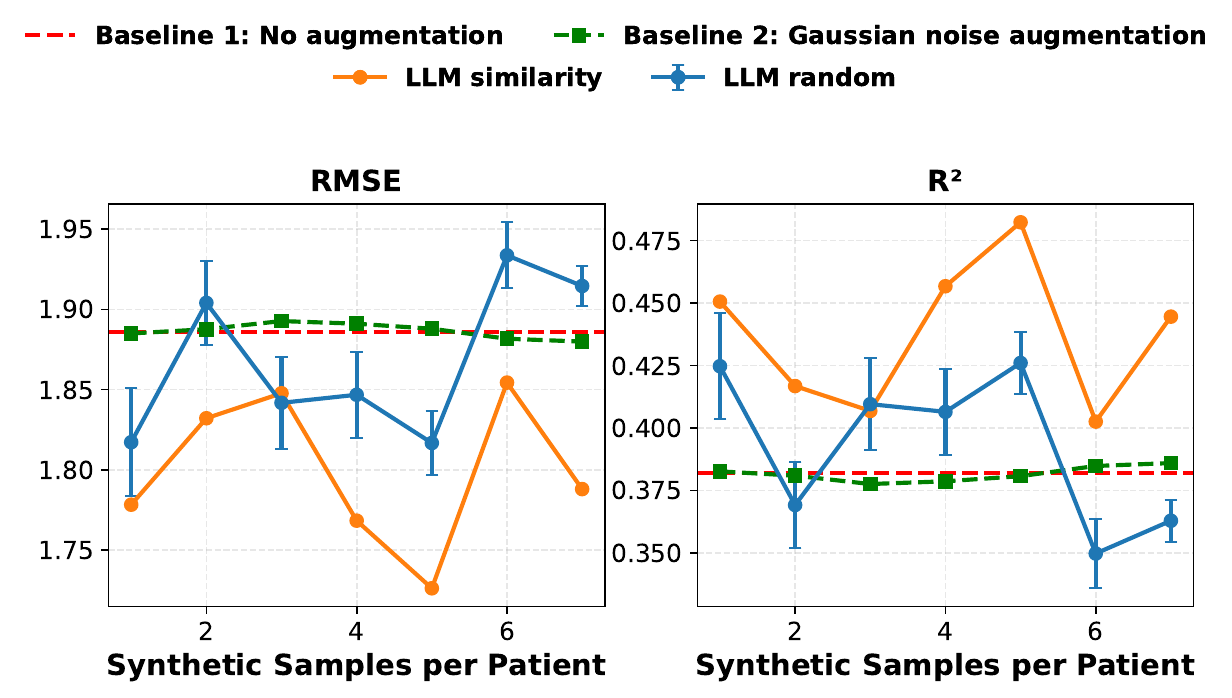}
\caption{RMSE and R² for different synthetic data augmentations.}
\label{fig:rmse_r2_comparison}
\end{figure}

Figure~\ref{fig:rmse_r2_comparison}\footnote{As MAE trends similarly to RMSE for both LLM-driven approaches, it is omitted for brevity.} shows the evolution of RMSE and R² as a function of the number of synthetic samples generated per patient, for several data augmentation strategies.

As a first observation, all augmentation methods improve performance compared to the baseline without augmentation, with lower RMSE and higher R². This confirms the relevance of synthetic data in low-data settings.

Building on this, LLM-driven approaches consistently outperform classical Gaussian noise augmentation. In particular, the LLM similarity-based strategy achieves the best performance across practically all configurations, suggesting that semantically coherent and clinically relevant samples are more effective than simple statistical perturbations.

Examining the effect of the augmentation scale, performance generally improves as more synthetic samples are added, although diminishing returns appear beyond a certain point (5 synthetic samples). This indicates that a limited number of high-quality synthetic samples may be sufficient to capture most of the performance gains.

Within the LLM-based methods, the random generation strategy provides substantial improvements over the baselines but remains consistently inferior to the similarity-based approach. Its slightly higher variability further highlights the importance of semantic control during synthetic data generation.

In contrast, Gaussian noise augmentation with $\sigma = 0.02$ yields only modest gains in RMSE and R², and slightly degrades MAE compared to the baseline, underscoring its limited ability to model the underlying structure and complexity of the data compared to LLM-based approaches.

Beyond raw performance, the similarity-based strategy exhibits an optimal point around five synthetic samples per patient, with a minimum RMSE of 1.7261 and a maximum R² of 0.4824, suggesting that performance can degrade beyond this level and that fewer, well-targeted synthetic samples may be more effective than larger volumes of less relevant data.

Finally, the parallel evolution of RMSE and R² indicates that improvements reflect both reduced error and better variance explanation. The absence of divergence between these metrics suggests that augmentation enhances generalization rather than inducing overfitting. 

\subsection{Critical Validation: True vs. Predicted Analysis}

Since our dataset is heavily dominated by healthy subjects (HDS 29–30), evaluating the model with global metrics alone can be misleading: a low overall MAE could simply indicate that the model predicts the mean score for all patients. To demonstrate the clinical value of our augmentation strategies, we therefore focus on the model's ability to predict low HDS scores (22–27), which represent the minority group and are critical for early detection of cognitive decline.

\begin{figure}[tb]
\centering
\includegraphics[width=0.7\linewidth]{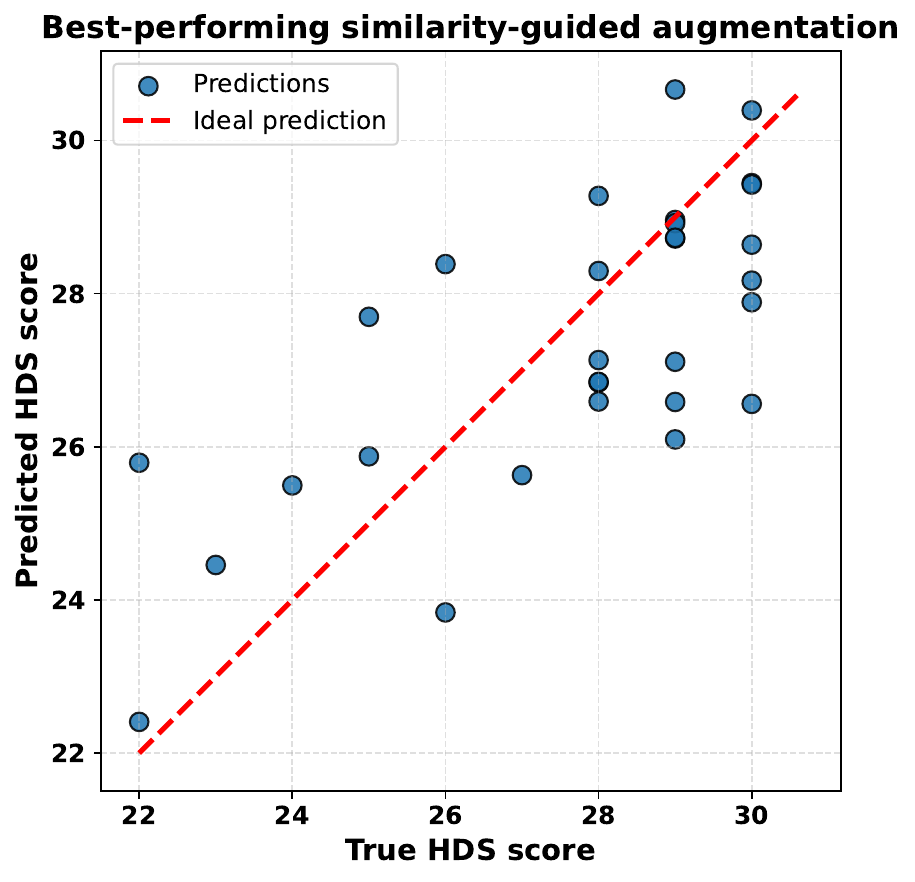}
\caption{True vs. predicted HDS scores for the best-performing similarity-based model.}
\label{fig:true_vs_predicted}
\end{figure}

We first visualize the model's predictions using a scatter plot of true versus predicted HDS scores for the similarity-based augmentation strategy with up to five synthetic samples per patient (Figure~\ref{fig:true_vs_predicted}). Points lying close to the diagonal indicate accurate predictions across both low and high score ranges. This visual check suggests that semantically guided synthetic data helps the model generalize beyond the majority high-score group, rather than simply regressing to the mean.

To quantify this observation, we perform a stratified evaluation by calculating the MAE separately for the minority (HDS 22–27) and majority (HDS 28–30) groups, as summarized in Table~\ref{tab:mae_stratified}. This allows us to directly assess whether augmentation strategies improve predictions for the clinically relevant minority group without compromising performance for the majority group.

\begin{table}[tb]
\centering
\begin{tabular}{ccc}
\toprule
\textbf{Method} & \textbf{Low Group} & \textbf{High Group} \\
 & (HDS 22–27) & (HDS 28–30)\\
\midrule
Baseline 1 & 2.381 & 1.236 \\
Baseline 2 & 2.378 & 1.255 \\
Proposed & 1.849 & 1.237 \\
\bottomrule
\end{tabular}
\caption{Stratified MAE for three augmentation approaches: Baseline 1 = no augmentation, Baseline 2 = Gaussian noise augmentation, and Proposed = similarity-guided LLM augmentation. Results are reported for the Low Group (HDS 22–27) and High Group (HDS 28–30).}
\label{tab:mae_stratified}
\end{table}

The results show that similarity-based augmentation substantially reduces MAE in the minority group while maintaining strong performance for the majority group. In contrast, Gaussian noise augmentation provides only modest improvements for the minority group and slightly increases MAE for the majority group, highlighting the superiority of semantically guided synthetic data in addressing class imbalance.

In summary, these analyses confirm that similarity-guided synthetic data not only improves overall metrics but, more importantly, enhances prediction accuracy for the clinically relevant minority group, demonstrating its potential utility for early detection scenarios.

\subsection{Analysis of Effective Styles}

\begin{figure}[t]
\centering
\includegraphics[width=\linewidth]{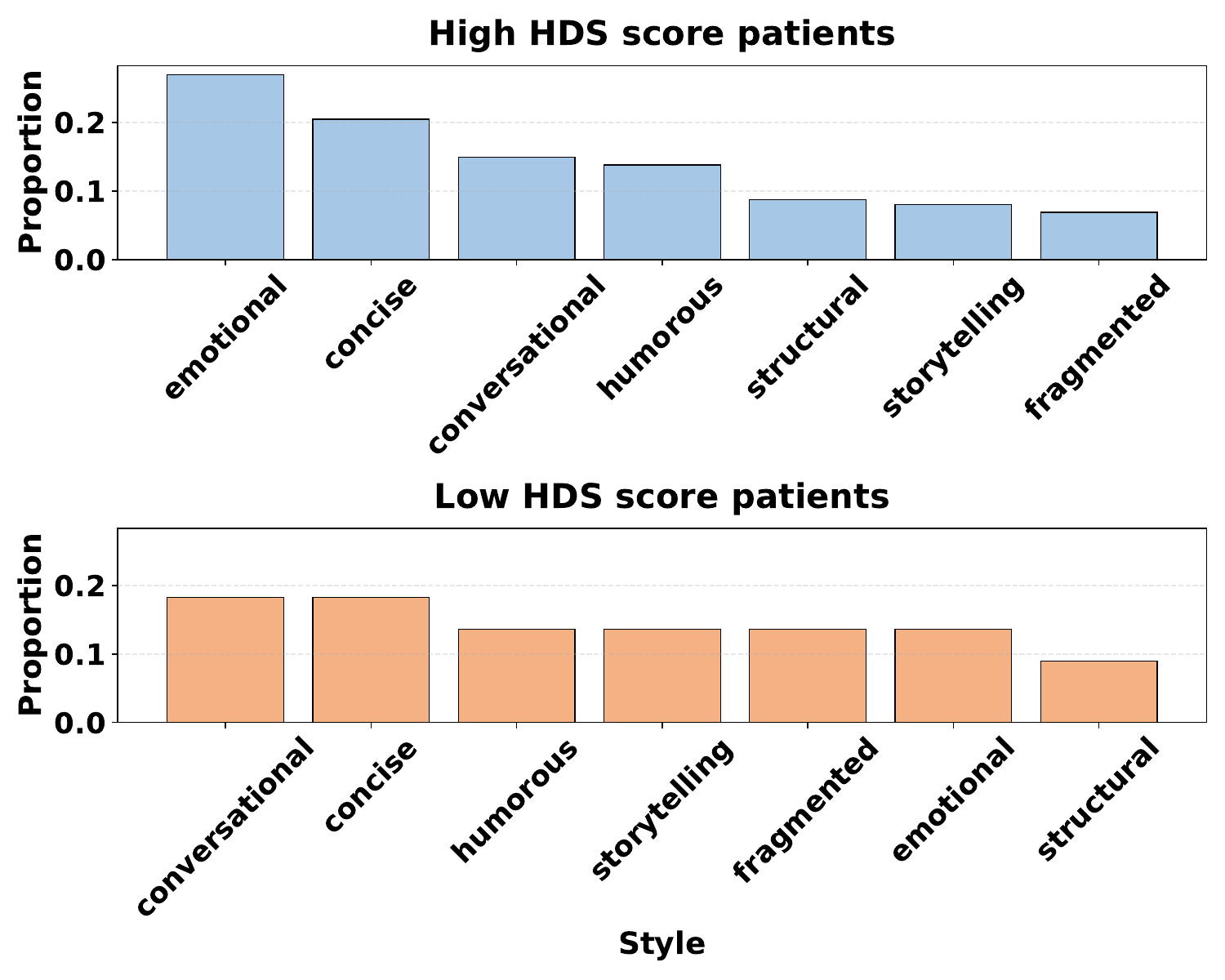}
\caption{Distribution of selected linguistic styles for patients with low and high HDS scores. Low and high HDS groups were defined arbitrarily for the purpose of this analysis.}
\label{fig:style_distribution}
\end{figure}

We analyzed the distribution of linguistic styles used by the best-performing similarity-guided augmentation strategy (five synthetic monologues) for patients with low and high HDS scores, as shown in Figure~\ref{fig:style_distribution}. For this analysis, we arbitrarily defined two HDS score groups: low scores (22–25) and high scores (26–30).

For low HDS score patients, the distribution of styles is relatively balanced, with several styles appearing at similar frequencies. This suggests that no specific style is inherently favored for these patients and that the selection is largely influenced by the similarity between the original and synthetic embeddings. One possible explanation is that, although a score of 22 indicates some cognitive decline relative to a perfect score of 30, the overall range of 22–30 still corresponds to patients without significant cognitive impairment, which may explain the lack of a clear stylistic trend. 

In contrast, high HDS score patients show a more skewed distribution, with certain styles amplified, indicating that when the original embeddings are more distinct or consistent, the augmentation strategy tends to select styles that closely align with them.

Overall, these observations suggest that the success of the similarity-guided augmentation depends primarily on embedding-based semantic proximity rather than any inherent preference for specific styles. Low-score patients reflect the variability of the original data, whereas high-score patients benefit from more consistent style alignment, demonstrating that prompt instructions and style variants interact with the underlying semantic structure of the data.

\section{Limitations}

Despite the promising results of similarity-guided class-balanced augmentation, several limitations of the current approach should be acknowledged.

First, the dataset exhibits severe class imbalances. Some HDS score classes, such as 23, 24, or 27, include only a single patient. Under the LOOCV evaluation scheme, holding out such individuals can result in training folds with very few or no examples from certain classes, limiting the model's ability to learn relevant patterns. The removal of synthetic monologues corresponding to the test patient, necessary to prevent data leakage, further reduces minority class representation during training.

Second, the selection of linguistic styles and the generation process constrain diversity. Identifying styles that are clinically relevant for early dementia detection remains an open question, and synthetic monologues are generated conditionally from the original patient responses, which may not fully address class imbalance or capture the full variability of speech patterns.

Third, the evaluation procedure itself has inherent limitations. While LOOCV is appropriate for the current small dataset, it is computationally intensive and may not scale efficiently to larger cohorts, limiting the generalizability of this protocol.

Finally, the current approach assumes that GPT-5 can generate synthetic monologues consistent with the patient’s cognitive state. This hypothesis carries some uncertainty, as the model's outputs are not guaranteed to fully reflect subtle variations in cognitive performance.

\section{Future Work}

While our current approach focuses on modifying the style of narratives while keeping their content fixed, an important next step would be to explore content modification while preserving the original cognitive style. Such experiments could help disentangle the contributions of narrative structure versus lexical style in predicting cognitive decline, shedding light on which aspects of language are most informative.

In parallel, a thorough evaluation against other established data augmentation methods is crucial to validate the advantages of our approach. Future comparisons could include simple paraphrasing techniques such as EDA or back-translation, as well as other LLM-based augmentation frameworks. These comparisons will allow us to situate our cross-modal augmentation framework within the broader landscape of text augmentation strategies and provide a more robust demonstration of its effectiveness.

Finally, extending the generation depth to larger values, would allow the study of long-term stylistic drift and cumulative effects across generations. While our current experiments were limited to 8 synthetic monologues due to hardware constraints, the observed results already highlight the strong potential of this iterative augmentation paradigm.

Together, these directions, namely content focused augmentation and systematic comparison with alternative methods, define a clear roadmap for extending and rigorously evaluating our framework, ultimately advancing the study of language based cognitive assessment.

\section{Conclusion}

In this work, we introduced a novel LLM-driven cross-modal-inspired data augmentation framework for cognitive score prediction from spontaneous speech. By leveraging written narratives as semantic anchors, our method generates synthetic oral-style monologues that preserve content while introducing stylistic variability, addressing the dual challenges of limited data and class imbalance in clinical datasets. Experimental results demonstrate that similarity-guided selection of synthetic samples not only improves overall regression metrics but, crucially, enhances prediction accuracy for the clinically important minority group, highlighting the potential of semantically coherent synthetic data for early detection of cognitive decline.

Beyond its immediate empirical benefits, our approach provides a principled framework for responsible synthetic data generation in clinical NLP, emphasizing semantic fidelity and cognitively informed style variation. These findings underscore the promise of large language models as powerful tools for data-efficient clinical assessment, paving the way for future research on content-focused augmentation, systematic comparisons with alternative methods, and ultimately, broader deployment in real-world cognitive screening applications.

By combining methodological rigor, clinical relevance, and practical scalability, this study contributes to the emerging paradigm of AI-driven cognitive assessment and sets a foundation for future advancements in leveraging language as a sensitive biomarker of neurodegeneration.

\section{References}
\addcontentsline{toc}{section}{References}
\bibliographystyle{lrec2026-natbib}
\bibliography{references}
\appendix
\section*{Appendix A: Prompt Template}
\addcontentsline{toc}{section}{Appendix A: Prompt Template and API Configuration}
\label{app:prompt}

This appendix provides the exact prompt template used to
generate the synthetic oral monologues described in Section~\ref{sec:4}.
The generation logic, style rationale, HDS-aware conditioning, and filtering
strategies are detailed in the main text and are not repeated here.

\subsection{Prompt Template}

Listing~\ref{lst:prompt} reproduces the \texttt{user} message sent to GPT-5
for every (patient, style) pair. The four placeholders are substituted at
runtime: \texttt{\{written\_text\}} with the patient's written narrative,
\texttt{\{score\}} with the HDS score, and \texttt{\{style\_name\}} /
\texttt{\{style\_description\}} with one row of Table~\ref{tab:styles}.
Each call also includes the fixed \texttt{system} message:
\textit{``You generate natural Japanese spoken monologues.''}

\begin{lstlisting}[
    breaklines=true,
    caption={Exact prompt template sent to GPT-5.
             Braces denote runtime placeholders.},
    label={lst:prompt}]
Convert the written Japanese text into an ORAL monologue.
Do NOT copy sentences. Keep only the meaning.

Written text:
---
{written_text}
---

Target style: {style_name}
Style description: {style_description}

Adapt fluency based on Hasegawa Dementia Scale score = {score}:
- 28-30: fluent, well-structured, richer vocabulary
- 25-27: mild hesitations, slightly simpler sentences
- 22-24: slower rhythm, simpler phrasing, occasional hesitations,
         slightly vague or less detailed expression

Constraints:
- Produce a monologue between 150 and 1300 Japanese characters.
- Add natural speech markers (e.g., etto, sono, ano)
- Avoid near-duplicates
- Output ONLY the monologue.
\end{lstlisting}

\subsection{Style Description Strings}

Table~\ref{tab:styles} lists the verbatim strings passed as
\texttt{\{style\_description\}} for each of the seven styles discussed in
Section~4.3.
\begin{table}[h]
\centering
\footnotesize
\begin{tabular}{@{}lp{5.8cm}@{}}
\toprule
\texttt{style\_name} & \texttt{style\_description} \\
\midrule
conversational         & Fluent conversational retelling, natural and relaxed tone. \\
emotional              & More emotional and reflective, slight introspection. \\
fragmented             & Spoken style with short segments, occasional breaks in rhythm, but still clear and coherent. \\
storytelling           & Chronological story-like narration with light details. \\
concise                & Simplified, concise spoken Japanese with short sentences. \\
humorous               & Slightly humorous or playful, with light jokes or witty expressions. \\
structural\_paraphrase & Preserves meaning and information density while reorganising sentence structure using similar vocabulary. \\
\bottomrule
\end{tabular}
\caption{Verbatim style description strings injected into the prompt template (Listing~\ref{lst:prompt}).}
\label{tab:styles}
\end{table}

\section*{Appendix B: Step-by-Step Generation Example}
\addcontentsline{toc}{section}{Appendix B: Step-by-Step Generation Example}
\label{app:example}

To illustrate how the pipeline transforms patient-produced material into
synthetic oral monologues, we trace one complete example from the original
written narrative through prompt instantiation to the final generated output.
All Japanese texts are presented here in English translation for readability.
The example uses a participant with an HDS score of 24 and the
\textit{emotional} style.

\setcounter{subsection}{0}  
\subsection{Step 1: Original Written Narrative (Patient Input)}

The following is an English translation of the handwritten response provided
by the participant in response to the clinical prompt \textit{``Please tell us
about the last good thing that happened to you.''}

\begin{quote}
\small\itshape
The other day, my grandchild came to visit. We went to a nearby park together
and ate a packed lunch. The weather was very nice and the cherry blossoms were
blooming beautifully. I was very happy watching my grandchild running around
energetically. When I bought ice cream on the way home, my grandchild was very
pleased.
\end{quote}

\noindent This written text is structured, grammatically complete, and contains
detailed descriptions, typical characteristics of the written modality in the
corpus.

\subsection{Step 2: Original Oral Transcription (Prediction Target)}

The following is an English translation of the transcription of the same
participant's spontaneous oral narrative recorded during the clinical session.
This is the actual data used as input for HDS score prediction.

\begin{quote}
\small\itshape
Um, the other day, you know, my grandchild came. Yeah, and then, uh, we went
to the park, and ate lunch. The weather was, you know, really nice. The cherry
blossoms were blooming too. Watching my grandchild running around, I just, you
know, felt happy. Oh, on the way home, I bought ice cream and my grandchild was
pleased. Yeah.
\end{quote}

\noindent Compared to the written version, the oral transcription is less
structured, contains hesitation markers (\textit{um, uh, you know, yeah}),
and uses simpler sentence constructions, reflecting natural spontaneous speech.

\subsection{Step 3: Instantiated Prompt}

The prompt template (Listing~\ref{lst:prompt}) is filled with the written
narrative from Step~1, the participant's HDS score (24), and the
\textit{emotional} style specification from Table~\ref{tab:styles}.
Listing~\ref{lst:instantiated} shows the resulting prompt as sent to GPT-5.

\begin{lstlisting}[
    breaklines=true,
    caption={Instantiated prompt for the example participant (HDS\,=\,24,
             emotional style). The written text placeholder is filled
             with the original Japanese narrative (English translation
             shown in Step~1).},
    label={lst:instantiated}]
Convert the written Japanese text into an ORAL monologue.
Do NOT copy sentences. Keep only the meaning.

Written text:
---
[Original Japanese written narrative inserted here]
---

Target style: emotional
Style description: More emotional and reflective, slight introspection.

Adapt fluency based on Hasegawa Dementia Scale score = 24:
- 28-30: fluent, well-structured, richer vocabulary
- 25-27: mild hesitations, slightly simpler sentences
- 22-24: slower rhythm, simpler phrasing, occasional hesitations,
         slightly vague or less detailed expression

Constraints:
- Produce a monologue between 150 and 1300 Japanese characters.
- Add natural speech markers (e.g., etto, sono, ano)
- Avoid near-duplicates
- Output ONLY the monologue.
\end{lstlisting}

\subsection{Step 4: Generated Synthetic Monologue (GPT-5 Output)}

The following is an English translation of the synthetic oral monologue
returned by GPT-5 in response to the instantiated prompt above:

\begin{quote}
\small\itshape
You know... the other day, my grandchild came to visit. Thinking about it
now, I really was so happy. We went to a nearby park together, and had lunch
there. The cherry blossoms were blooming, and, well, the weather was just so
nice... Watching my grandchild running around so full of energy, I thought,
ah, this is what makes life good, you know. And then, uh, on the way home, I
bought some ice cream. Seeing my grandchild's face light up like that...
yeah, that really made me happy. I still think about it.
\end{quote}

\noindent Compared to the original oral transcription (Step~2), the 
synthetic monologue preserves the same factual content and hesitation 
markers, while introducing a more reflective and introspective tone, 
consistent with the \textit{emotional} style specification.

\end{document}